\documentclass[letterpaper, 10 pt, journal, twoside]{IEEEtran} 

\IEEEoverridecommandlockouts                              

\usepackage{cite}
\usepackage{url}
\usepackage{amsmath}
\usepackage{amssymb}
\usepackage{caption}
\usepackage{pifont}
\usepackage{graphicx} 
\usepackage{subcaption}
\usepackage{stfloats}
\title{\LARGE \bf
GC-GAT: Multimodal Vehicular Trajectory Prediction using Graph Goal Conditioning and Cross-context Attention 
}

\author{Mahir Gulzar$^{1}$, Yar Muhammad$^{2}$, and Naveed Muhammad$^{1}$}

\usepackage{fancyhdr}
\usepackage{lipsum} 

\usepackage[colorlinks=true, linkcolor=blue, citecolor=blue, urlcolor=blue]{hyperref}

\begin{document}

\maketitle
\thispagestyle{fancy}

\renewcommand{\thefootnote}{}
\footnotetext{%


This publication has been financed by the European Social Fund via the "ICT programme" measure, and by collaboration project LLTAT21278 with Bolt Technologies.

$^{1}$Mahir G. and Naveed M. are associated with the Institute of Computer Science, University of Tartu, Tartu 51009, Estonia. {\tt\small FirstName.LastName@ut.ee}

$^{2}$Yar M. is associated with the University of Hertfordshire, Hatfield, AL10 9AB, United Kingdom. {\tt\small y.muhammad@herts.ac.uk}

}
\renewcommand{\thefootnote}{\arabic{footnote}}

\begin{abstract}

Predicting future trajectories of surrounding vehicles heavily relies on what contextual information is given to a motion prediction model. The context itself can be static (lanes, regulatory elements, etc) or dynamic (traffic participants). This paper presents a lane graph-based motion prediction model that first predicts graph-based goal proposals and later fuses them with cross attention over multiple contextual elements. We follow the famous encoder-interactor-decoder architecture where the encoder encodes scene context using lightweight Gated Recurrent Units, the interactor applies cross-context attention over encoded scene features and graph goal proposals, and the decoder regresses multimodal trajectories via Laplacian Mixture Density Network from the aggregated encodings. Using cross-attention over graph-based goal proposals gives robust trajectory estimates since the model learns to attend to future goal-relevant scene elements for the intended agent. We evaluate our work on nuScenes motion prediction dataset, achieving state-of-the-art results.

\end{abstract}

\section{INTRODUCTION}\label{sec:introduction}

Autonomy, in the context of autonomous vehicles (AV)s, can be defined as a vehicle's ability to operate without a human's intervention \cite{1}. Autonomous driving, when achieved, promises many benefits, including (i) safer roads, as most of the accidents are caused by human error, (ii) reduction in emissions as a result of more efficient driving, (iii) mobility for all, e.g., children and the elderly, (iv) and ease, comfort, and freeing up time \cite{2}. 

Arguably \cite{3}, one of the most challenging sub-problems in autonomous driving is an estimation of others' intentions in traffic. For us humans, it's instinctive to predict the actions of other agents (be it other human-driven vehicles, pedestrians, or bi-cyclists, etc.) a few seconds into the future. This, however, is a nontrivial task for an autonomous vehicle. Being able to predict other agents' future actions is important for the purpose of safety and efficiency. While an autonomous vehicle that drives extremely slowly might be able to avoid any collisions with other road agents without predicting their future actions (and just performing simple obstacle detection), such a vehicle wouldn't be very useful. Therefore, for increased efficiency while ensuring safety (in other words, for driving safely at reasonable speeds), autonomous vehicles need to be able to reliably predict the future actions of agents around them.

There are several methods to represent the future motion of an agent. \cite{4} proposes a taxonomy of motion prediction where predicted output is categorized into intention, uni-modal/multi-modal trajectory, and map occupancies. Here, the intention is too general to be used in AV planning, whereas map occupancies are usually compute-intensive. The literature we see today mostly revolves around predicting the future trajectories of intended agents. We see this in recent competitions in motion prediction proposed by various organizations \cite{5,6,7}. A database consisting of a high-definition (HD) map along with agent histories is provided to the participants. The prediction results are expected to be in the form of multimodal trajectory.

Many solutions have been proposed by researchers to address motion prediction on existing datasets. Early works include \cite{8}, which predicts multi-modal trajectory by rasterizing HD map into multiple semantic layers. A vanilla Convolutional neural network (CNN) extracts relevant scene features from the rasters here. Similarly, \cite{9,10} uses rasters to encode the scene's contextual information. Since motion prediction is a crucial step of the autonomy pipeline, these approaches cannot be used in real-time due to their compute intensity. Additionally, when encoded, rasterized input suffers from a large neural receptive field, i.e., the CNN feature maps often hallucinate, thus resulting in incorrect predictions. More recent works exploit the structured data of HD maps. Here, VectorNet \cite{11} was among the pioneers that directly used a sequence of vectors to represent HD-map data. \cite{12} used a similar structured representation to represent lanes of HD maps as graph's nodes and edges. In such regard, graph-based networks are used to encode map information. These works include LaneRCNN \cite{13}, which applies lane graph convolution to encode map lanes that are near traffic agents; Lane-based Trajectory Prediction (LTP) \cite{14}, which uses Graph Neural Network (GNN) and Transformer to encode map and traffic data and \cite{15}, which first encodes environment context using GRUs (Gated Recurrent Units) and later processes that information using GAT (Graph Attention Network) \cite{16}. In \cite{15}, authors also used lane graphs to train a graph traversal policy.

While works discussed above encode both temporal and spatial scene context using graph-based networks and heuristics, they are less attentive toward target-centric goal nodes. We argue that an additional step of encoding goal proposals over lane graphs and applying cross-attention over them will capture future dependencies of target agent with respect to lane graphs. To this end, we propose \textbf{GC-GAT}: Graph Conditioned Goal ATtennion: a multimodal trajectory prediction network conditioned on agent-centric graph goals. Our model is inspired by the encoder-aggregator-decoder architecture of \cite{15}, i.e., we first encode the scene context (lanes, agents) using GRUs and process lane encodings with GAT. In aggregator, instead of graph traversal policy, we apply simple cross attentions over lanes and dynamic agents. A $k$ moded goal proposal is predicted for the target agent, and probabilities for these modes are extracted. In the decoding step, in addition to the latent variable noise $z$, we incorporate a Laplacian MDN (Mixture Density Network) proposed by \cite{17} to predict context-aware future trajectories.

Our main contributions can be summarized as follows:

\begin{enumerate}
  \item We propose a novel multimodal trajectory prediction network conditioned on agent-centric graph goals. Our network applies cross-attention over proposed graph goals to capture better future dependencies for the intended agent. 
  \item We evaluate the performance of our model on the NuScenes \cite{5} public dataset, achieving state-of-the-art results.
\end{enumerate}

\section{Related Work}

Vehicular trajectory prediction has been addressed using various modeling techniques. Some of the well-known works can be categorized as following:

\textbf{Scene encoding:}
Encoding scene-relevant information is crucial for trajectory prediction. This includes encoding static (map elements) and dynamic (traffic participants) context. Static scene information is mostly available in the form of HD-map containing lanes and regulatory elements whereas dynamic context is overlayed on top HD-map with timestamps. As discussed in Section-\ref{sec:introduction}, the trends in encoding scene features have shifted from rasterized inputs \cite{8,9,10} to polyline (geometrical) vector encoding \cite{13,14,15,16,18}. This is mainly due to the fact that like any other module of autonomy pipeline, prediction is required in real-time, i.e. a performance friendly prediction model reduces bottlenecks in the decision making pipeline.


\textbf{Interaction via attention}: Since their emergence, transformers \cite{19} have been in limelight and are extensively applied in different subdomains of computer science. One such application is capturing interactions between different input features for behavior modeling. In graph-based methods (discussed above), we see different use cases of multi-headed attention, e.g., capturing dependencies and interactions between lane to lane, lane to agent, agent to agent, etc.
For example, \cite{20} encodes a variable number of neighboring agents with graph nodes using an attention network. \cite{21} uses stacked transformer layers that aggregate
the contextual information on the fixed trajectory proposals. \cite{22} employs separate attention heads to capture possible interaction between the target agent and the context features. An interaction transformer is proposed in \cite{23} that jointly models interactions among all scene actors. Our work uses multi-headed attention (MHT) to capture interactions between lanes, neighbors, target agents, and graph goal encodings.

\textbf{Multimodal predictions:} Keeping in view the uncertainty in an agent's possible future, it is preferable to have multimodal predictions to get more than one future hypothesis. Multimodal outputs can be deterministic or directly sampled from a probability distribution. In encoder-decoder architectures, the decoder part often directly regresses multimodal predictions \cite{24, 25}; on the downside, they suffer moderately from mode collapse. Other works use latent variable often termed $z$ sampled from a simple distribution to a predicted trajectory \cite{15, 18}. We mostly see latent variable models in Generative-adversarial Networks (GANs) \cite{28} often trained with Variational Autoencoder (VAE) \cite{29} and Conditional Variational Autoencoder (CVAE) \cite{30}. These works include \cite{31, 32, 33, 34, 35}. Since ground truth only consists of one trajectory per agent, most of the above methods suffer from mode collapse if training is not handled efficiently. One solution to this is using Laplacian Mixture Density Networks (MDNs) in the regression step \cite{17, 26, 27}. Our work in this paper, in addition to $z$ latent vector, uses Laplacian MDN in the decoding step. This is followed by a winner-takes-all strategy loss that handles mode collapse much more efficiently rather than directly decoding the trajectories \cite{15, 17, 36, 37}.

\begin{figure*}[htbp]
\vspace{0.2in}
\centering
\includegraphics[width=1 \textwidth]
{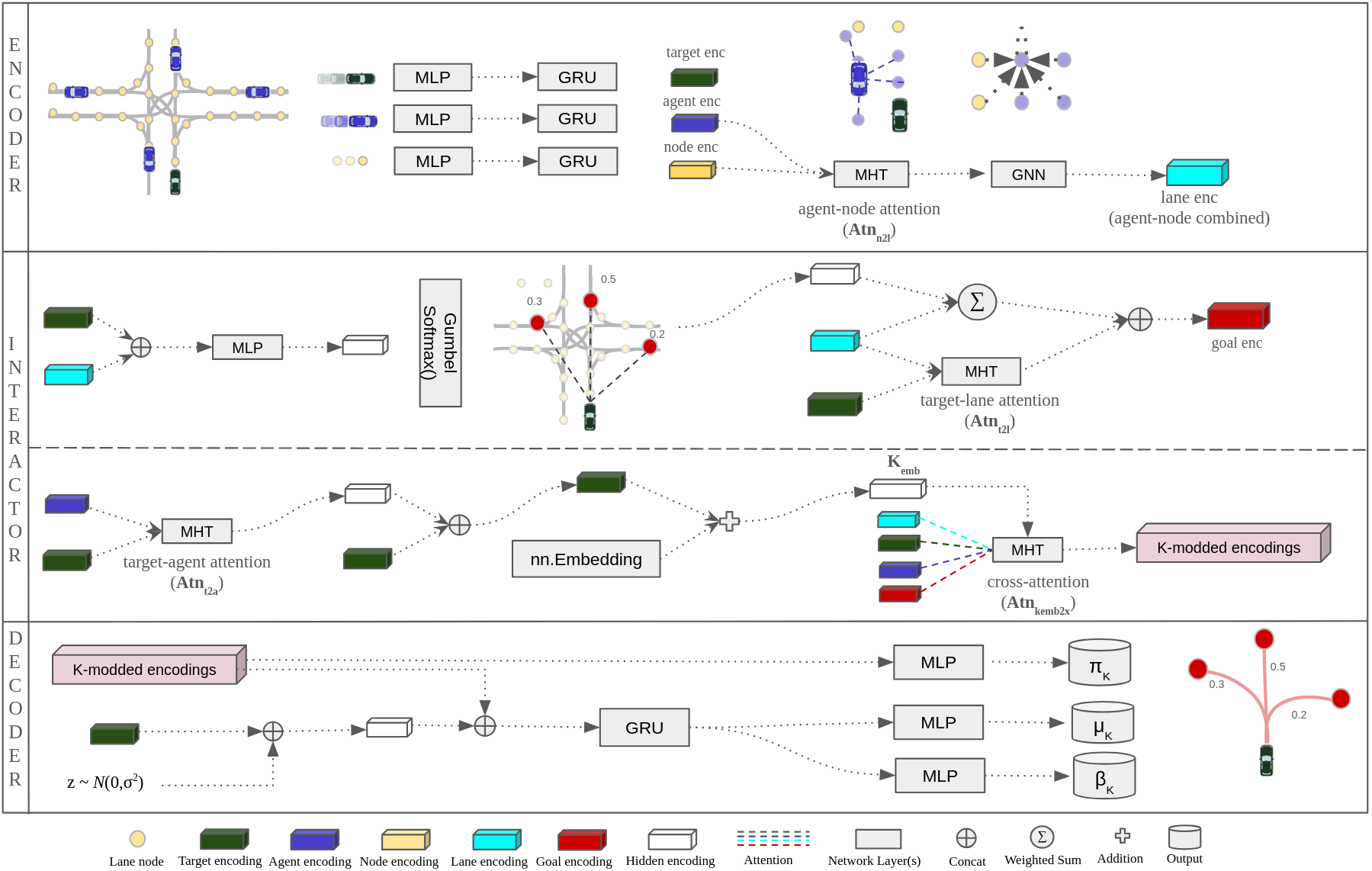}
\caption{Model architecture of GC-GAT. The model is divided into three parts. \textbf{Encoder:} Encodes static and dynamic context and applies agent-node attention followed by graph attention using GNN layers. \textbf{Interactor:} Computes goal encodings in the first step using Multi-Layer Perceptron (MLP). A weighted sum of output of MLP is concatenated with target-lane attention to get the final goal encodings. In the second step, the interactor applies target-agent attention followed by an aggregation with Embedding layer to calculate K-modded embeddings which are later fed to MHT layers for cross-attention of K-modded embedding with different types of encodings to produce K-modded encodings. \textbf{Decoder:} Firstly, applies a simple MLP to K-moded encodings to extract mode probabilities $\pi_k$ which is one of the outputs of the model. Secondly, it adds Gaussian noise to target encodings and concatenates it K-moded encodings which are fed to a GRU followed by two identical MLPs that extract location $\mu_k$ and scale $b_k$ parameters of Laplace mixture density network which are the remaining two outputs of the model along with $\pi_k$. It is to be noted that the data flows from left to right in each row, and architecture should read in the same manner, going from top to bottom.}
\label{fig:gc_gat_model_architecture}
\end{figure*}

\section{Problem Statement}

Multimodal trajectory prediction for a given agent can be formulated as follows:

At any given instance, a scene consists of multiple traffic agents represented by their past, present, and future spatial coordinates. A top-down birds-eye-view (BEV) containing 2D coordinates (can be agent-centric) 
denoted by $\textbf{X}_{t_h:t}^i$ representing a sequence of agent $i$'s history from timestep $t_h$ consisting of $h$ history steps in the past up to current timestep $t$ is input to the model. Whereas $\textbf{Y}_{t+1:t_f}^i$ a sequence of agent $i$'s future from timestep $t+1$ to $t_f$ for $f$ number of future steps represents the ground-truth for the agent $i$ where $i \in \mathbb{N}_1$. Let $\textbf{C}$ be the static scene context vector:

\begin{equation}
\hat{\textbf{Y}}_{t+1:t_f,k
}^i = f\left( \textbf{X}{t_h
}^i, \textbf{C} \right)
\label{eq:multimodal_prediction_funct}
\end{equation}

Here, $k$ represents number of modes and $k>1$. The optimization is done via a loss function that measures the difference between the predicted values $\hat{\textbf{Y}}_{t+1:t_f, k}^i$ and the ground truth $\textbf{Y}_{t+1:t_f}^i$.

\begin{equation}
\mathcal{L} = \text{loss}\left( \hat{\textbf{Y}}_{t+1:t_f
, k}^i, \textbf{Y}_{t+1:t_f
}^i \right)
\label{eq:generic_loss_funct}
\end{equation}

Since $k$ modes represent a distribution, at least one mode should closely approximate the ground truth for successful training.

\section{Methodology}

This section describes main structure of our proposed model, including input and output to the proposed model followed by the model architecture.

\subsection{Input}
We refer to \cite{15, 18} for input representation. Specifically, the inputs come from two different types of context, i.e., dynamic and static. The contextual input to the model is target agent-centric, i.e., since this work proposes a single-agent trajectory predictor, all relevant information given to the model is relative to one target agent in the scene at any given instance.  The model inputs are formalized as follows:
\subsubsection{\textbf{Dynamic Context}}
 A variable set of agents in the scene $\mathcal{A} = \{A_0,...,A_i\}$ where $i \in \mathbb{N}_1$ represents total number of agents in the scene (should be at least one). An agent's historical data consists of $A_i = (x_{i},y_{i}, v_i, a_i, w_i, \textit{I}) \in \mathbb{R}^6$ where $x_i,y_i$ are 2D top-down spatial coordinates of the agent $v_i,a_i,w_i$ represent speed, acceleration, and yaw\_rate. Here, $\textit{I}$ is a boolean flag identifying the agent type as 1 if the underlying agent is pedestrian and 0 otherwise.
\subsubsection{\textbf{Static Context}}
The HD map is represented by a directed lane-node graph $G(V,E)$. Here, ${V} = \{V_1,...,V_v\},  v\in \mathbb{N}$ is a set of lane graph nodes where each node consists of equidistant segments of lane center lines. For consistency, the number of poses inside node segments are of fixed length. A $vth$ node is represented by $V_v = (x_{n}^v, y_{n}^v, \theta_{n}^v, \textit{I}_{n}^v)$ where $x_{n}^v, y_{n}^v, \theta_{n}^v$ are spatial coordinates with yaw for the $nth$ pose of $vth$'s segment. Similar to dynamic context, $\textit{I}_{n}^v$ is a boolean flag identifying whether lane segment's pose is inside a stopline or crosswalk. Unlike \cite{15}, we don't exploit the graph traversals in this paper. Instead, we only treat graph nodes as feature vectors, so edges $E$ and their types are irrelevant.

\subsection{Output}

The output trajectory of the model is almost exactly similar to $\hat{\textbf{Y}}_{t+1:t_f, k}^i$ from (\ref{eq:multimodal_prediction_funct}). We can elaborate it as a sequence of spatial coordinates, i.e. $\hat{\textbf{Y}}_{t, k}^i = \{(x_{t,k}^i,y_{t,k}^i |t = 1,...,t_{f}\}$ where $(x_{t,k}^i,y_{t,k}^i)$ is the $ith$ trajectory point of $kth$ hypothesis at $t$ timestep. Additionally, since we intend to use winner-takes-all training strategy, a vector $P_k$ is also predicted by the model representing the probability score for each mode.

\addtolength{\topmargin}{0.2in}
\subsection{Model}

Fig.~\ref{fig:gc_gat_model_architecture} shows an illustration of our proposed model. Our model follows the famous encoder-interactor-decoder architecture \cite{15, 17, 18, 36} with cherry-picked layers of existing works. The encoder module is essentially encoding features using GRU's and processing them with global graph attention. The encoded features are passed over to the interactor, which further processes the information using different multiheaded attention layers, each responsible for capturing interactions between dynamic and static context. Apart from this, the iteractor module also predicts graph goal proposals, which are later fused with the attention layers. As a final step, the decoder regresses multi-mode trajectory using Laplace MDN. Here, we also leverage latent variable noise $z$ for longitudinal variability. The final results consisting of multi-modal trajectory and mode scores are then used to calculate the overall loss. Further details of each submodule are as follows:

\subsubsection{\textbf{Encoder}}

Since both static and dynamic context consists of sequences of points and we intend to learn both spatial and temporal features of the input, The input features are first processed independently by simple MLPs for each type of context as shown in the encoder part of Fig.~\ref{fig:gc_gat_model_architecture}. The MLP embeddings are then handed over to independent GRUs to extract temporal features. GRUs are preferred here because they are lightweight recurrent modules with faster training speeds and less memory usage than their counterpart Long short-term memory (LSTM)s . The dynamic context is further divided into two in the encoder for better understandability. These are target-agent and surrounding agent encodings. The above encoding steps are shown below:
\[
h_{target} = GRU(MLP(A_{target}))
\]
\[
h_{nbr} = GRU(MLP(A_{nbr}))
\]
\[
h_{lane} = GRU(MLP(V_{lane}))
\]
Here, $h_{target}$, $h_{nbr}$, and $h_{lane}$ are independent hidden encoded states of the target agent, neighboring agents, and lane nodes, respectively. Once we obtain these hidden states, the encoder applies a MHT layer between $h_{nbr}$ and $h_{lane}$. This captures the first cross-interaction between neighboring agents and the lane nodes.
\[
Atn_{n2l} = MultiHead(Q_{nbr},K_{lane},V_{lane})
\]
\[
h_{lane} = h_{lane} \oplus Atn_{n2l}
\]
\[
h_{lane} = GNN(h_{lane})
\]
Here, the query is neighbors hidden encoding, whereas keys and values come from lane encodings. To exploit the graph structure, in the final step of the encoder, the previous attention vector is concatenated with lane encodings and processed by a simple GNN consisting of GAT layers. This constitutes the encoder's output for the next submodule.

\subsubsection{\textbf{Interactor}}

In contrast to \cite{36}, the interactor consists of three parts. The hidden features of target-agent, neighbouring agents and lane are first fed to MHT layers to capture cross-attention between the dynamic and static context. These attentions are represented as:
\[
Atn_{t2l} = MultiHead(Q_{target},K_{lane},V_{lane})
\]
\[
Atn_{t2a} = MultiHead(Q_{target},K_{nbr},V_{nbr})
\]
Here, $Atn_{t2l}$ is target-agent to lane graph attention and $Atn_{t2a}$ is target-agent to neighbour agents attention. These attention outputs are consumed by two different branches of the model. 


In the second step, a target-centric goal predictor computes goal probabilities for the lane node graph. Initially, a simple MLP is applied over the concatenated hidden encodings of the target agent and lane graph. The output encodings are then processed by the Gumbel-Softmax to obtain soft samples.
\[
\mathbf{g}_{\text{enc}} = MLP(h_{\text{target}} \oplus h_{\text{lane}})
\]
\[
\mathbf{g}_{\text{weighted}} = GumbelSoftmax(\mathbf{g}_{\text{enc}}, \tau)
\]

Here, \( \mathbf{g}_{\text{weighted}} \) represents soft samples derived from the categorical distribution defined by \( \mathbf{g}_{\text{enc}} \). The temperature parameter \( \tau \) controls the smoothness of the distribution during the Gumbel-Softmax operation. 

Finally, we extract the goal encodings by computing a weighted sum of the lane node encodings \( h_{\text{lane}} \) and \( \mathbf{g}_{\text{weighted}} \).
\[
h_{\text{goal}} = \sum_{i} g_{\text{weighted}}^{(i)} h_{\text{lane}}^{(i)}
\]
Where \( h_{goal} \) represents the goal node encodings. Finally, the goal node encodings are concatenated with target-centric lane graph attention we computed previously.
\[
h_{goal} = Atn_{t2l} \oplus h_{goal}
\]
In the third step of the interactor, the target encodings are concatenated with $Atn_{t2a}$. Afterward, $k_{emb}$ embeddings are added and fused with four different attention layers, each capturing $k$ mode-oriented attention over the hidden encodings. 
\[
h_{target} = h_{target} \oplus Atn_{t2l} \oplus Atn_{t2a}
\]
\[
k_{emb} = nn.Embedding(K, D),
\]
\[
k_{emb} = k_{emb} + h_{target}
\]
\[
Atn_{k_{enc}2x} = MultiHead(Q_{k_{emb}},K_{x},V_{x})
\]
\[
k_{enc} = norm(Atn_{k_{emb}2x})
\]
Here, \(k_{emb}\) represents a trainable embedding matrix initialized randomly, with \(K\) being the number of trajectory modes and \(D\) the dimensionality of each embedding. $Atn_{k_{enc}2x}$ represent MHT layers for capturing k modded attention between $k_{emb}$ and $x$, where $x$ are hidden encodings consisting of $h_{target}, h_{lane}, h_{nbr}$ and $h_{goal}$. It is to be noted that our model uses three types of aggregations i.e. concat for retaining distinct features, simple addition for merged unification, and weighted sum to learn weights of influence on final tensor.

\begin{table*}[!b]
    \centering
    \caption{The benchmarking results on \textit{nuScenes} test set on public leaderboard}
    \small 
        \begin{tabular}{cccccc}
            \hline
            \noalign{\vskip 2mm} 
            \textbf{Works} & 
            \textbf{MinADE$_5$} $\downarrow$ & 
            \textbf{MinADE$_{10}$} $\downarrow$ & 
            \textbf{MissRate$_5$} $\downarrow$ & 
            \textbf{MissRate$_{10}$} $\downarrow$ & 
            \textbf{OffRoadRate} $\downarrow$ \\
            \noalign{\vskip 2mm} 
            \hline
            \noalign{\vskip 0.7mm} 
            MultiPath \cite{25} & 2.32  & 1.96 & - & - & - \\
            \noalign{\vskip 0.7mm}
            CoverNet \cite{10} & 1.96 & 1.48 & 0.67 & - & - \\
            \noalign{\vskip 0.7mm}
            Trajectron++ \cite{9} & 1.88 & 1.51 & 0.70 & 0.57 & 0.13 \\
            \noalign{\vskip 0.7mm}
            AgentFormer \cite{39} & 1.86 & 1.45 & - & - & - \\
            \noalign{\vskip 0.7mm}
            MHA-JAM \cite{40} & 1.81 & 1.24 & 0.59 & 0.46 & 0.07 \\
            \noalign{\vskip 0.7mm}
            CXX \cite{41} & 1.63 & 1.29 & 0.69 & 0.60 & 0.08 \\
            \noalign{\vskip 0.7mm}
            LaPred \cite{42} & 1.47 & 1.12 & 0.53 & 0.46 & 0.09 \\
            \noalign{\vskip 0.7mm}
            GOHOME \cite{43} & 1.42 & 1.15 & 0.57 & 0.47 & 0.04 \\
            \noalign{\vskip 0.7mm}
            Autobot \cite{44} & 1.37 & 1.03 & 0.62 & 0.44 & 0.02 \\
            \noalign{\vskip 0.7mm}
            THOMAS \cite{45} & 1.33 & 1.04 & 0.55 & 0.42 & 0.03 \\
            \noalign{\vskip 0.7mm}
            PGP \cite{15} & 1.27 & 0.94 & 0.52 & 0.34 & 0.03\\
            \noalign{\vskip 0.7mm}
            XHGP \cite{18} & 1.28  & 0.95 & 0.53 & 0.34 & 0.03 \\
            \noalign{\vskip 0.7mm}
            MacFormer \cite{46} & 1.21 & \textbf{0.89} & 0.57 & \textbf{0.33} & 0.02 \\
            \noalign{\vskip 0.7mm}
            LAformer \cite{17} & 1.19 & 1.19 & \textbf{0.48 }& 0.48 & \textbf{0.02} \\
            \noalign{\vskip 0.7mm}
            GC-GAT (Ours) & \textbf{1.19} & 1.06 & 0.52 & 0.49 & 0.03 \\
            \noalign{\vskip 0.7mm}
            \hline
        \end{tabular}
    \label{tab:benchmarks}
\end{table*}

\subsubsection{\textbf{Decoder}}
The decoder is inspired from \cite{15}, \cite{17} and \cite{36}. Here, at the very first step, we use an MLP to extract k-mode probabilities $\pi_K$, where $\pi_K$ is merely a Laplacian mixing coefficient.
\[
\pi_K = MLP(k_{enc})
\]
Afterward, a Gaussian noise $z$ is added to $h_{target}$ encodings. This helps to capture longitudinal variability in the predictions. The Gaussian noise-based target encodings are then aggregated with k-modded encodings $k_{enc}$ we got from the interactor. It is noted that $k_{enc}$ are query-based encodings that contain attention over all hidden encodings, whereas $h_{target}$ are only target-agent's hidden encodings.

\[
h_{target} = h_{target} \oplus z \sim \mathcal{N}(0, \sigma^2)
\]

The aggregated encodings $h_{target}$ $\oplus$ $k_{enc}$ are then processed in a GRU cell for decoding the temporal dimension of the output as suggested by \cite{17}.

\[
out_{temp} = GRU(h_{target} \oplus k_{enc})
\]

Since Laplace distribution is presented by location $\mu$ and scale $b$ parameters. $f(x \mid \mu, b) = \frac{1}{2b} \exp\left( -\frac{|x - \mu|}{b} \right)$
 We employ two MLPs to output these values for K modes. This makes it a Laplacian Mixture Density Network such that for K modes $f(x \mid \{\pi_K, \mu_K, b_K\}) = \sum_{k=1}^{K} \pi_k \frac{1}{2b_k} \exp\left( -\frac{|x - \mu_k|}{b_k} \right)
$ where $\pi_K$ was already predicted earlier and $\{\mu_K, b_K\}$ are predicted by the two MLPs.
\[
\mu_K = MLP(out_{temp})
\]
\[
b_K = MLP(out_{temp})
\]
The use of Laplacian MDNs gives us the leverage of incorporating different types of loss functions. We use Laplace negative log-likelihood and cross-entropy losses for trajectory regression and mode classification. Alongside this, we use classical minADE loss between predicted trajectories and the ground-truth. Eq. (\ref{eq:generic_loss_funct}) changes as follows:
\begin{equation}
\mathcal{L} = L_{reg} + L_{cls} + L_{ADE}
\end{equation}
Here,
\[
L_{reg} = \frac{1}{t_f} \sum_{t=1}^{t_f} - \log P(\textbf{Y}_{t} \mid \mu_t^{m^{*}}, b_t^{m^{*}})
\]
where $m^*$ represents the prediction mode with minimum $\textit{L}_2$ error compared to ground-truth $\textbf{Y}_{t}$.
\[
L_{cls} = \sum_{k=1}^{K} (-\pi_k \log (\hat{\pi_k}))
\]
where $\hat{\pi_k}$ is predicted mode probability and $\pi_k$ is soft target probability [17].
\[
L_{ADE} = \arg\min_{k} \left( \frac{1}{T} \sum_{t=1}^{T} \| \hat{\textbf{Y}}_t^{(k)} - \textbf{Y}_{t} \| \right)
\]
where $\hat{\textbf{Y}}_t^{(k)}$ represents the predicted mode with the minimum average displacement error.

\begin{table*}[!t]
    \vspace{0.2in}
    \centering
    \caption{Ablation of interactor module}
    \small 
        \begin{tabular}{cccccc}
            \hline
            \noalign{\vskip 2mm} 
            \textbf{Goal Proposals} &
            \textbf{Cross Attention} &
            \textbf{MinADE$_5$} $\downarrow$ & 
            \textbf{MinADE$_{10}$} $\downarrow$ & 
            \textbf{MissRate$_5$} $\downarrow$ & 
            \textbf{MissRate$_{10}$} $\downarrow$ \\
            \noalign{\vskip 2mm} 
            \hline
            \noalign{\vskip 0.7mm} 
             &  & 1.70 & 1.56 & 0.71 & 0.70 \\
            \ding{51} &  & 1.35  & 1.22 & 0.62 & 0.59 \\
             & \ding{51} & 1.21  & 1.07 & 0.52 & 0.49 \\
            \ding{51} & \ding{51} & \textbf{1.19}  & \textbf{1.06} & \textbf{0.52} & \textbf{0.49} \\
            \noalign{\vskip 0.7mm} 
            \hline
        \end{tabular}
    \label{tab:interactor_ablation}
\end{table*}

\addtolength{\topmargin}{-0.2in}
\section{Experiments}
\subsection{Dataset}
Our work is evaluated on the nuScenes \cite{5} dataset. The dataset consists of more than 1000 scenes, and each scene is a sequence of 20 seconds. A scene comprises HD-maps and different types of traffic participants covering many urban road scenarios. Additionally, even though nuScenes provide spatial and temporal data of different dynamic agents, it only evaluates the vehicular agents for the final benchmarks.
\subsection{Metrics}
For evaluation, we employ the standard nuScenes evaluation metrics \cite{38}. These metrics include minimum average displacement error or point-wise $L2$ error between prediction and ground-truth over top K modes ($MinADE_K$), final displacement error or $L2$ distance between final points of prediction and ground-truth trajectory over top K modes ($MinFDE_K$), miss rate over K modes ($MissRate_K,2$) i.e., if max point-wise distance between predicted mode and ground-truth is greater than 2 meters it is considered a miss and is evaluated over K most likely predictions and finally, offroad rate, that measures the fraction of how much of the predicted trajectory is going beyond the drivable area.

\subsection{Results}

\begin{figure*}[!t]
  \centering

  \begin{subfigure}[b]{0.22\textwidth}
    \includegraphics[width=\linewidth, height=1.05in]{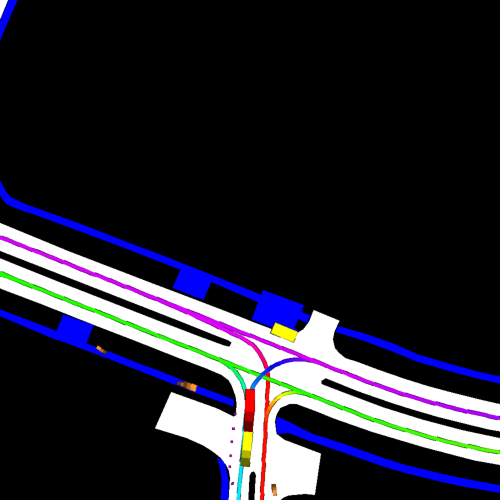}
  \end{subfigure}
  \begin{subfigure}[b]{0.22\textwidth}
    \includegraphics[width=\linewidth, height=1.05in]{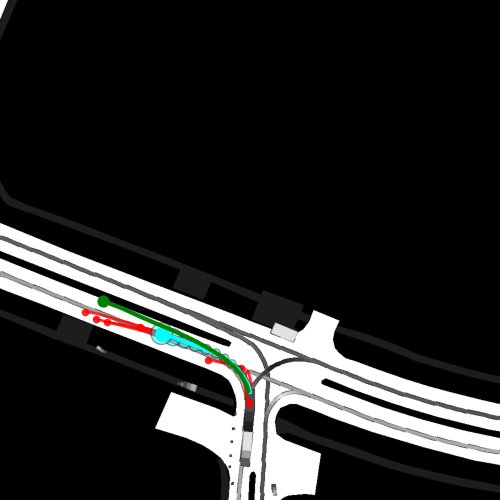}
  \end{subfigure}
    \hspace{0.005\textwidth}
  \begin{subfigure}[b]{0.22\textwidth}
    \includegraphics[width=\linewidth, height=1.05in]{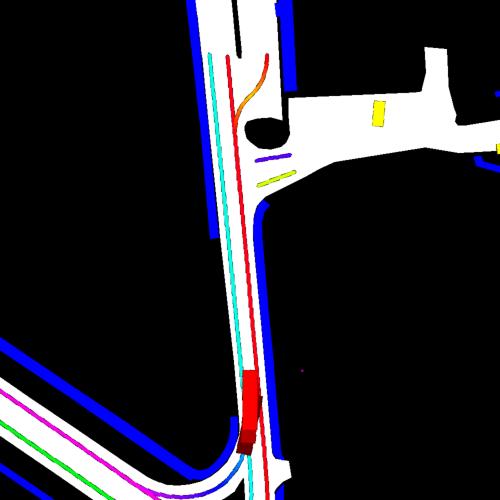}
  \end{subfigure}
  \begin{subfigure}[b]{0.22\textwidth}
    \includegraphics[width=\linewidth, height=1.05in]{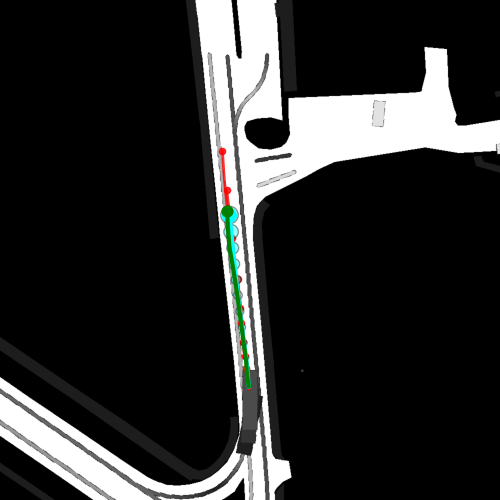}
  \end{subfigure}

 \vspace{0.05in} 
 
    \begin{subfigure}[b]{0.22\textwidth}
    \includegraphics[width=\linewidth, height=1.05in]{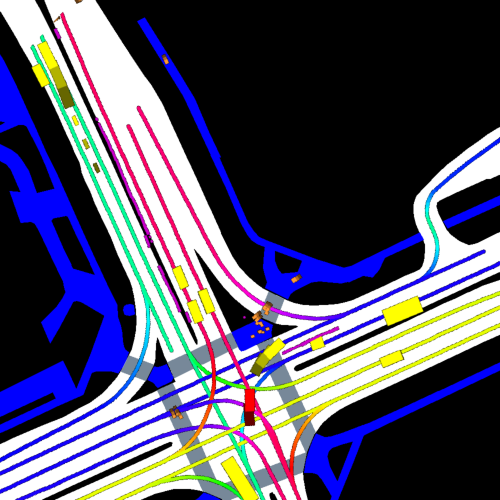}
  \end{subfigure}
  \begin{subfigure}[b]{0.22\textwidth}
    \includegraphics[width=\linewidth, height=1.05in]{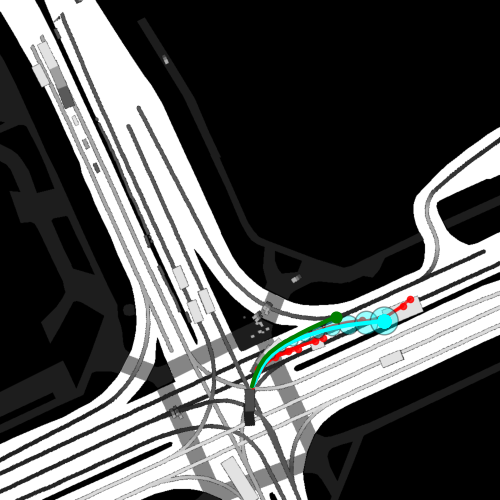}
  \end{subfigure}
  \hspace{0.005\textwidth}
  \begin{subfigure}[b]{0.22\textwidth}
    \includegraphics[width=\linewidth, height=1.05in]{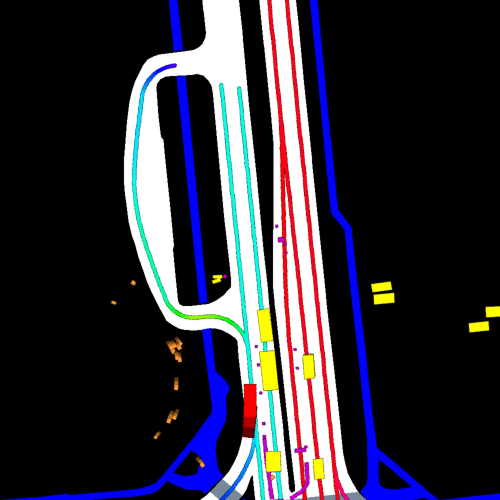}
  \end{subfigure}
  \begin{subfigure}[b]{0.22\textwidth}
    \includegraphics[width=\linewidth, height=1.05in]{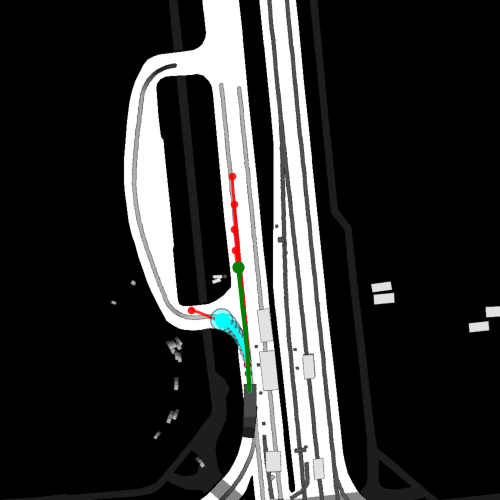}
  \end{subfigure}

 \vspace{0.05in} 

    \begin{subfigure}[b]{0.22\textwidth}
    \includegraphics[width=\linewidth, height=1.05in]{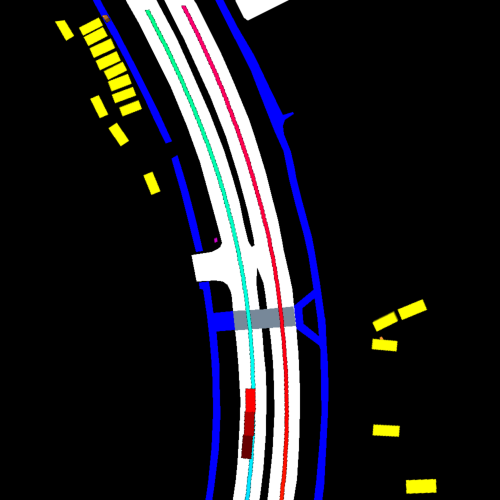}
  \end{subfigure}
  \begin{subfigure}[b]{0.22\textwidth}
    \includegraphics[width=\linewidth, height=1.05in]{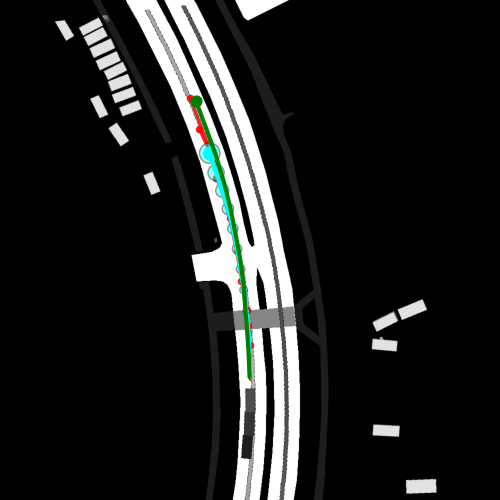}
  \end{subfigure}
  \hspace{0.005\textwidth}
  \begin{subfigure}[b]{0.22\textwidth}
    \includegraphics[width=\linewidth, height=1.05in]{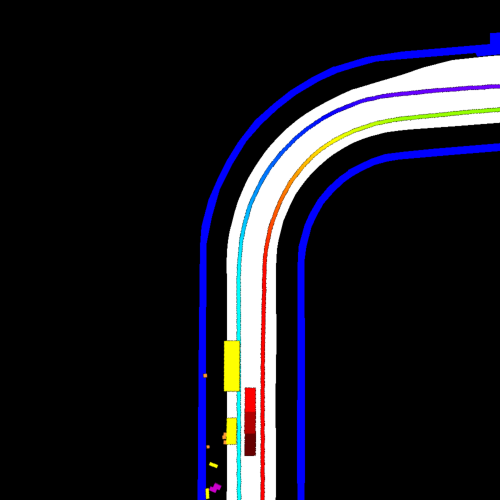}
  \end{subfigure}
  \begin{subfigure}[b]{0.22\textwidth}
    \includegraphics[width=\linewidth, height=1.05in]{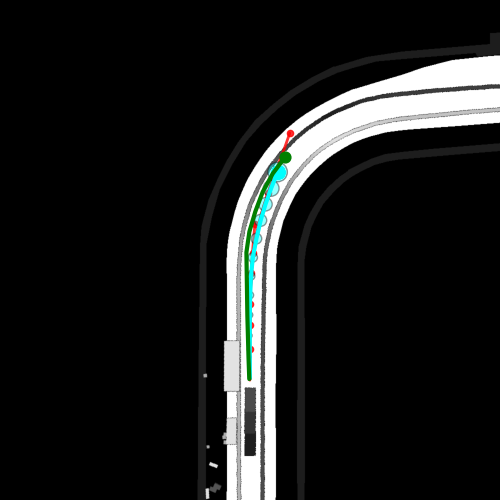}
  \end{subfigure}

  \vspace{0.13in}
  \centering

  \begin{subfigure}[b]{0.22\textwidth}
    \includegraphics[width=\linewidth, height=1.05in]{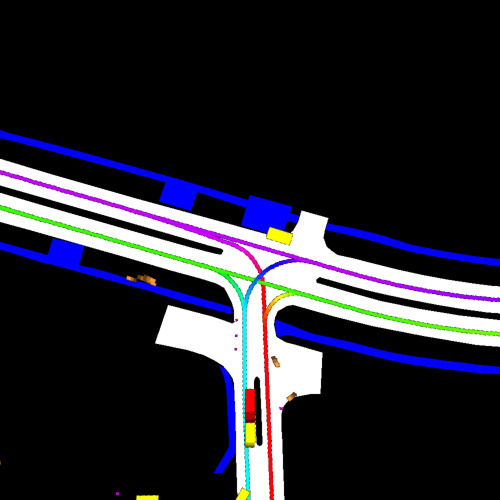}
  \end{subfigure}
  \begin{subfigure}[b]{0.22\textwidth}
    \includegraphics[width=\linewidth, height=1.05in]{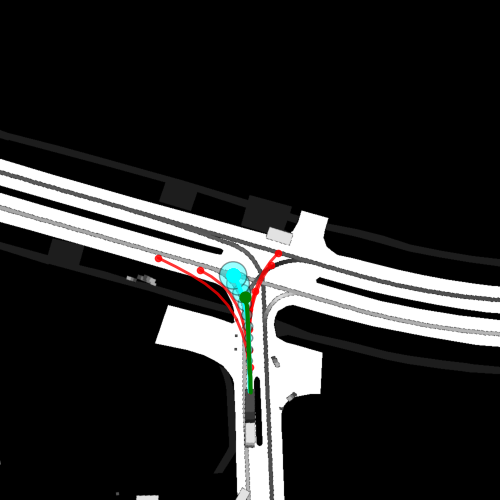}
  \end{subfigure}
    \hspace{0.005\textwidth}
  \begin{subfigure}[b]{0.22\textwidth}
    \includegraphics[width=\linewidth, height=1.05in]{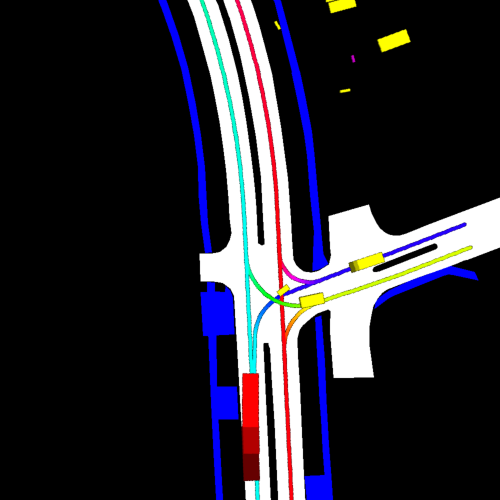}
  \end{subfigure}
  \begin{subfigure}[b]{0.22\textwidth}
    \includegraphics[width=\linewidth, height=1.05in]{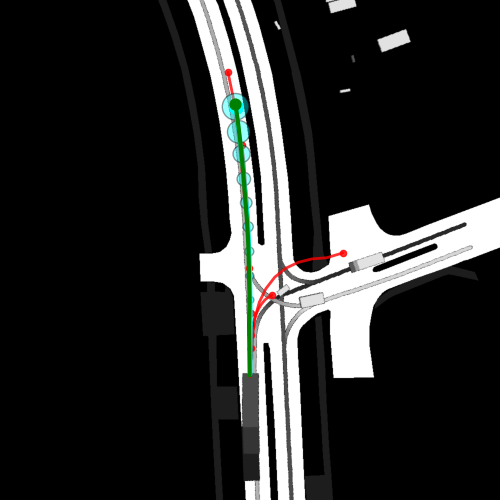}
  \end{subfigure}

 \vspace{0.05in} 
 
    \begin{subfigure}[b]{0.22\textwidth}
    \includegraphics[width=\linewidth, height=1.05in]{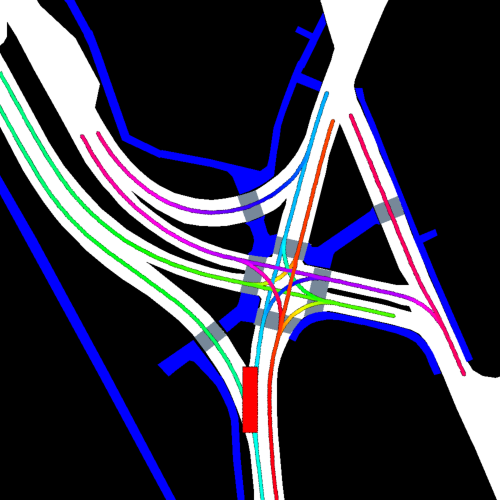}
  \end{subfigure}
  \begin{subfigure}[b]{0.22\textwidth}
    \includegraphics[width=\linewidth, height=1.05in]{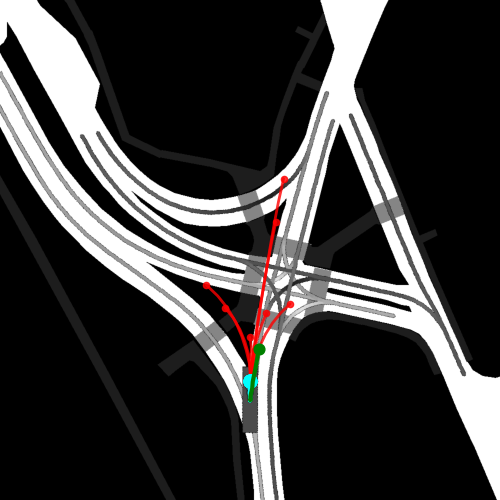}
  \end{subfigure}
  \hspace{0.005\textwidth}
  \begin{subfigure}[b]{0.22\textwidth}
    \includegraphics[width=\linewidth, height=1.05in]{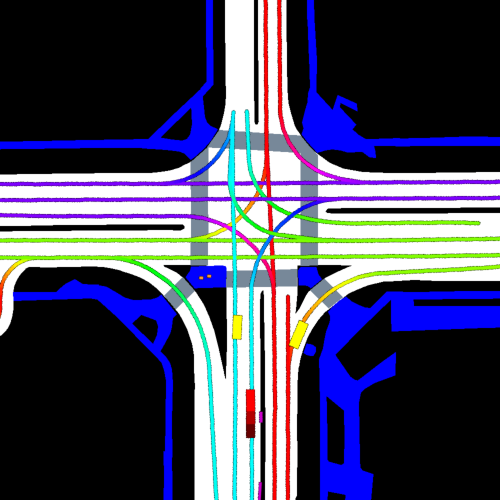}
  \end{subfigure}
  \begin{subfigure}[b]{0.22\textwidth}
    \includegraphics[width=\linewidth, height=1.05in]{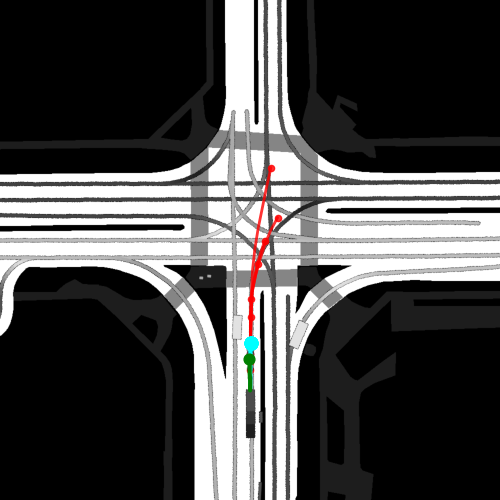}
  \end{subfigure}

 \vspace{0.05in} 

  \caption{Qualitative results of GC-GAT on nuScenes[5] test set. The BEV map shows the target vehicle (red bounding box) and other traffic participants (yellow and orange boxes). Adjacent greyscale raster displays ground-truth trajectory (green) and predicted trajectories (red and cyan, with cyan representing the most likely trajectory). Uncertainty of the Laplace MDN is visualized as cyan circles (only for the most likely trajectory), with larger circles indicating higher uncertainty. Notably, uncertainty increases with the prediction horizon.}\label{fig:qualitative_results}
\end{figure*}

The benchmarking results of our model are reported in Table 1. Our model achieved a score comparable to the current state-of-the-art on the nuScenes leaderboard. Regarding MinADE$_5$, our model gave equal performance to LAformer \cite{17} which is one the best ranking results on the leaderboard. For MinADE$_{10}$, our model outperformed LAformer but lacked in achieving the best scores. For MissRate$_5$, MissRate$_{10}$ and OffRoadRate our results are comparative to MacFormer\cite{46}, PGP\cite{15}, XHGP\cite{18} and THOMAS\cite{45}. Notebly, Autobot\cite{44}, GOHOME\cite{43}, LaPred\cite{42} and CXX\cite{41} were completely outperformed by GC-GAT.

For qualitative comparison, Fig.~\ref{fig:qualitative_results} shows positive examples along with some prominent limitations of our model with different scene settings on nuScenes test set. Here, the first and third columns show a rasterized HD map with traffic participants whereas columns two and four show prediction results from the perspective of the target agent. We showcase different types of road scenes, including intersections, T-junctions, left-right turns, and curves. The top three rows show positive results (reasonable trajectory modes) whereas the last two rows demonstrate the model's two most prominent generalizability limitations i.e. firstly, predicted trajectories with plausible goal estimates sometimes go through non-drivable islands (shown in the second-last row), secondly, the model sometimes predicts illogical goals i.e. goals over wrong lanes, resulting in illogical trajectory modes (shown in the last row). In general, the predictions don't suffer mode-collapse in all of the shown scenes, i.e., diverse outputs are predicted with respect to road lanes.

\subsection{Ablation}

In this section, we analyze the effects of replacing different network layers and operations from the interactor module. The interactor is chosen here because the current architecture of encoder and decoder has already been ablated in \cite{15} and \cite{17}. In such regard, Table.\ref{tab:interactor_ablation} shows a quantitative comparison of model's performance under different interactor settings. Having no cross-attention and goal proposals gave the worst performance as expected since no interactions were captured, and aggregated encodings were passed through to the decoder module. In the next step, we evaluated the model's performance by introducing goal proposals and their cross-attention with aggregated scene encodings. This significantly improved the benchmarking results. Similar to goal proposals, in the third step, we introduced cross-attention among target, surrounding, and node encodings (goal encodings were excluded here), which performed even better than simple goal proposals. The best results we got from the interactor are by fusing goal encodings and cross-attention with static and dynamic context.

\subsection{Discussion}

Since the study conducted in this work is done on a preprocessed, clean, publicly available dataset, our evaluations are confined to the dataset on which the model is trained. Most of the existing literature on prediction models has an inherent problem of adaptability, i.e., adapting and evaluating these models to practical autonomy pipelines suffers from performance disparity, also known as \textit{dynamics gap} \cite{47}. While GC-GAT's explicit evaluation on an autonomous driving stack isn't performed due to the scope of this study, it can be ported to any real-time autonomy pipeline since it has an average prediction time of approx. $11.6$ milliseconds (per batch prediction), when tested on an RTX 3070 mobile GPU with 8GB VRAM. For practical application, the autonomy pipeline should support the same input formulation on which the model is trained. A dynamic evaluation mechanism proposed in \cite{47} can also be employed for real-time model evaluation.

\section{Conclusions}

This work presented GC-GAT (Graph Conditioned Goal ATtention) model for predicting multi-modal trajectories. Our model is inspired by encoder-interactor-decoder architecture and is built on relevant existing works. The modal encodes scene information using light-weight GRU's and interactions are captured using multi-headed attention. Our model predicts goal proposals and applies cross-attention over encoded scene context and goal proposals, making it robust in capturing future goal-relevant dependencies. We benchmarked GC-GAT on the famous nuScenes dataset, where we received comparable scores with the current state-of-the-art, beating different baselines. Lastly, we discussed model's performance both quantitatively and qualitatively with addition of ablation study for better understandability of model's architecture and tuning. In conclusion, simple goal conditioning with graph cross-attention gave efficient results for modeling multimodal trajectories.



\bibliographystyle{ieeetr}

\addtolength{\topmargin}{0.15in}
\bibliography{root}

\end{document}